\documentclass{article}
\usepackage{ijcai25}

\usepackage{times}
\usepackage{soul}
\usepackage{url}
\usepackage[hidelinks]{hyperref}
\usepackage[utf8]{inputenc}
\usepackage[small]{caption}
\usepackage{graphicx}
\usepackage{amsmath}
\usepackage{amsthm}
\usepackage{booktabs}
\usepackage{algorithm}
\usepackage{algorithmic}
\usepackage[switch]{lineno}
\usepackage{subfigure}
\usepackage{amsfonts} 
\linenumbers

\newtheorem{defn}{Definition}
\newtheorem{thm}{Theorem}

\title{Model-Based Offline Reinforcement Learning with \\ Reliability-Guaranteed Sequence Modeling}

\author{
Shenghong He
\and
Chao Yu\and
Qian Lin\and
Yile Liang\and 
Donghui Li\And
Xuetao Ding\\
\affiliations
\emails
}

\begin{document}

\maketitle

\nolinenumbers
\begin{abstract}
Model-based offline reinforcement learning (MORL) aims to learn a policy by exploiting a dynamics model derived from an existing dataset. 
Applying conservative quantification to the dynamics model, most existing works on MORL generate trajectories that approximate the real data distribution to facilitate policy learning by using current information (e.g., the state and action at time step $t$).
However, these works neglect the impact of historical information on environmental dynamics, leading to the generation of unreliable trajectories that may not align with the real data distribution.
In this paper, we propose a new MORL algorithm \textbf{R}eliability-guaranteed \textbf{T}ransformer (RT), which can eliminate unreliable trajectories by calculating the cumulative reliability of the generated trajectory (i.e., using a weighted variational distance away from the real data).
Moreover, by sampling candidate actions with high rewards, RT can efficiently generate high-return trajectories from the existing offline data.
We theoretically prove the performance guarantees of RT in policy learning, and empirically demonstrate its effectiveness against state-of-the-art model-based methods  on several benchmark tasks.
\end{abstract}

\section{Introduction}

Model-based offline reinforcement learning (MORL)~\cite{DBLP:KidambiRNJ20} aims to learn a policy by leveraging a dynamics model derived from an existing dataset, without requiring interaction with the environment.
Recent approaches often leverage uncertainty quantification~\cite{DBLP:YuTYEZLFM20,DBLP:WagenmakerSJ23,DBLP:BhardwajRSAHCK24} and value function penalization~\cite{yu2021combo,DBLP:SunZJLYY23,DBLP:LuisBVB023} to learn the dynamics model, ensuring that the data generated based on current information (e.g., state and action) align with the real data distribution.

However, these methods implicitly assume that the current information fully provides all relevant information for data generation at the next time step. Thus, unreliable trajectories that may not exist in the real environment can be generated due to neglect in considering the impact of historical information on the environmental dynamics.
For instance, considering a robot performing cleaning tasks in a complex room,
at a historical time step $t-k$, the robot takes an action of moving a chair, causing a partial obstruction in the passage.
At the current time step $t$ , employing a generative model to generate trajectories could result in the robot passing through the chair due to the model's inability to remember the historical action of moving a chair.
This generated trajectory is unreliable because the robot cannot pass through the chair in the real environment.
Furthermore, by simply focusing on imitating the observed data transitions, existing methods generally fail to identify states and actions that lead to high cumulative returns, thus significantly impairing the learning performance.

To address the above issues, in this paper, we propose a novel method called \textbf{R}eliability-guaranteed \textbf{T}ransformer (RT).
Specifically, taking advantage of the sequence modeling capabilities of transformer, RT captures historical information between different positions in the input sequence to generate data (e.g., states, actions, and rewards).
Subsequently, RT computes the variation distance between the generated data and the real data to derive a reliability value of the generated data for each time step,
and then incorporates a truncation metric based on the cumulative reliability along the trajectory to adaptively determine the trajectory length, so as to mitigate the accumulation of errors and consequently enhances the overall reliability of the generated sequences.
Moreover, RT estimates the likelihood of future high returns for candidate actions and uses these estimations as generation conditions to generate high-return trajectories.
Our contributions are as follows:
\begin{itemize}



    \item We explore the role of leveraging historical information to generate reliable trajectories with high future returns to facilitate and stabilize policy learning in MORL.

    

    \item We propose RT that enhances trajectory reliability by incorporating cumulative reliability into sequence modeling and continuously selects actions with high rewards to generate high-return trajectories.

    \item Empirical results on various continuous control tasks demonstrate that our method  achieves superior learning  performance compared to existing baselines.
    
\end{itemize}

\section{Preliminaries}
\label{sec:preli}
\textbf{MDPs and offline RL}.
A Markov Decision Process (MDP) can be defined as $M=(S,A,R,P,\rho_0,\gamma)$, where $S$ is the state space, $A$ is the action space, $R(s, a)$ is the reward function, $P(s'|s,a)$ is the transition function, $\rho_0$ is the the initial state distribution, and $\gamma \in (0,1)$ is the discount factor.
A policy $\pi:S \times A \rightarrow [0,1]$ takes action $a$ at state $s$ with probability $\pi(a|s)$.
The goal of RL is to find the optimal policy $\pi^*$ that maximizes the expected return $\pi^* = \mathop{\text{argmax}}_\pi\limits \mathbb{E}_\pi[\sum_{t=0}^T \gamma^t R(s_t,a_t)]$.
In the offline RL setting, the agent only has access to a static dataset $ \mathcal{D}_{\text{env}} = \{ (s, a, r, s') \} $.
The agent's objective is to learn a policy  without interaction with the environment for any additional online exploration.
\\
\\
\textbf{Model-based offline  RL}.
Model-based offline RL methods use a dataset to learn the dynamics model $\hat{P}$, which is usually trained by maximum likelihood estimation: $\mathop{\min}_{\hat{P}}\limits \mathbb{E}_{(s,a,s') \sim D_{env}}[-\log \hat{P}(s'|s,a)]$.
Additionally, when the reward function is unknown, a model of the reward function $\hat{R}(s,a)$ can be trained.
Once the dynamics model is trained, samples generated by $\hat{P}$ are then placed into the model buffer $\mathcal{D}_{\text{model}}$, which is merged with the offline data $\mathcal{D}_{\text{env}} \cup \mathcal{D}_{\text{model}}$ to learn a policy.
In this work, we assume that the transition function of the environment is determined and propose RT to augment the offline data.

\section{Method}
The overall process of RT is described in Algorithm~\ref{alg:RT_algo}.
RT first learns a dynamics model from the offline dataset and establishes a threshold for the reliability of the offline data (lines 1 to 11). 
Next, RT randomly selects a trajectory from the offline dataset and generates a reverse trajectory from any state along that trajectory.
This generated trajectory is then combined with the original data to create a new dataset, which is used to train the model-free algorithm (lines 12 to 21).

\begin{algorithm}[tb]
\caption{The RT algorithm}
\label{alg:RT_algo}
\textbf{Input}: Offline dataset $D_{env}$, generation number H, iteration N, model-free offline RL algorithm (e.g., BCQ)\\
\textbf{Parameter}: Randomly initialize RT parameters $\theta$, $\psi$  and $\phi$,\\
\textbf{Output}: offline RL algorithm
\begin{algorithmic}[1] 
\STATE // Train RT
\FOR{i=0, $\cdots$, N}
\STATE Compute $\mathcal{L}_\tau$ using the dataset $D_{env}$
\STATE Updata RT network parameter $\theta$
\ENDFOR
\STATE // Train VAE to get the cumulative threshold $\alpha$
\FOR{for j=0, $\cdots$, N}
\STATE Compute $L(\psi,\phi)$ using offline dataset $D_{env}$
\STATE Update VAE network parameter $\psi$ and $\phi$
\ENDFOR
\STATE Use VAE to calculate the maximum distribution error of offline dataset to get a cumulative reliability threshold $\alpha$
\STATE //Generate trajectories
\STATE Initialize the buffer $D_{model}$
\FOR{i=0, $\cdots$, H}
\STATE Sample state $s_{t+1}$ from the $\tau$ in $D_{env}$
\STATE Generate backward trajectory \\ $\hat{\tau}=\{s_{t-k},a_{t-k},r_{t-k}\}_{k=0}^{t}$ using RT and reliability estimation.
\STATE Merge trajectories $\tau'=\hat{\tau}_{<T-t}+\tau_{\geq T-t}$
\STATE $D_{model} \leftarrow D_{model} \cup \tau'$
\ENDFOR
\STATE  Get composite dataset $D \leftarrow D_{env} \cup D_{model}$
\STATE Learn the model-free offline algorithm using $D$
\end{algorithmic}
\end{algorithm}

\subsection{Reliability-Guaranteed Sequence Generation}
Inspired by previous sequence representation learning~\cite{DBLP:DT}, we treat each input trajectory $\tau$ as a sequence and add reward-to-go $R_t=\sum^T_{h=t} \gamma^{h-t}r_h$ as an auxiliary information at each time step $t$, which acts as future heuristics for further generating data. 
Specifically, each trajectory $\tau=(s_1,a_1,r_1,R_1,\cdots,s_T,a_T,r_T,R_T)$ is used as an input sequence for RT, which is trained using the standard teacher-forcing method~\cite{DBLP:HeWWST22}, as expressed by: 
\begin{equation}
\begin{split}
     \log P_\theta(\tau_t|\tau_{<t}) &= \log P_\theta(s_t|\tau_{<t})+\log P_\theta(a_{t}|s_t,\tau_{<t}) \\
    & +\log P_\theta(r_t|a_{t},s_t,\tau_{<t}) \\
    &+\log P_\theta(R_t|r_t,a_t,s_t,\tau_{<t}),
\end{split}
\end{equation}
where $\tau_{<t}$ represents the trajectory from the initial state to time step $t-1$, $\theta$ is the parameter of RT, and $P_\theta$ is the induced conditional probability.
The training objective is to maximize:
\begin{equation}
\label{eq:RT_loss}
    \mathcal{L}_\tau=\sum_{t=1}^{T} \log P_\theta(\tau_{t} | \tau_{<t}).
\end{equation}

However, during the sequence generation, the model generation error at moment $t$ is continuously accumulated by the subsequent generation process, which may cause the generated data to deviate from the distribution of the original data.
In order to solve this issue, we propose a  reliability estimation mechanism, which automatically determines the generated trajectory lengths based on the cumulative reliability along the trajectory and incorporates these reliability values into the pessimistic MDP~\cite{DBLP:BlanchetLZZ23} to provide performance bounds for the learned policy.
We first define the cumulative reliability of trajectories and the truncation metric:
\begin{defn}[\textbf{Cumulative Reliability}]
    Given a trajectory $\tau$, the cumulative reliability along the trajectory at step $t$ is defined as:
    \begin{equation}
    \begin{split}
    &\Gamma(s_t,\tau_{<t})=\mathbb{D}(P(\cdot|s_i,\tau_{<i}),\hat{P}(\cdot|s_i,\tau_{<i})) \\
    &=\sum^t_{i=1} \frac{\exp(e_i)}{\sum_{j=1}^n \exp(e_j)} \cdot D_{IST}(P(\cdot|s_i,a_i),\hat{P}(\cdot|s_i,a_i)),
    \end{split}
    \end{equation}
\end{defn}

\noindent \textit{where $\hat{P}$ is the estimated transition probability of the environment after training, $P$ is the true dynamics, $e$ is an attention value, and $D_{IST}$ represents the total variation distance~\cite{DBLP:HoY10a} between two distributions $\hat{P}$ and $P$.}

\begin{defn}[\textbf{Truncation Metric}]
   Given a cumulative reliability $\Gamma$, the truncation metric at $t$ as:
    \begin{equation}
        U_t=\left \{ \begin{array}{lcl} 0 \quad (i.e., \ reliable) \ if \ \Gamma(s_t,\tau_{<t}) \leq \alpha \\ 1 \quad (i.e., unreliable) \ otherwise \end{array} \right. ,
    \end{equation}
\end{defn}
\noindent where \textit{$\alpha$ represents the cumulative threshold for generating reliable trajectories}.

$\Gamma$ provides a metric of the cumulative reliability of the trajectories generated by the model, while $U$ defines the truncation point of the generated trajectory. 
To directly quantify the impact of generation errors on policy learning, inspired by previous works~\cite{DBLP:KidambiRNJ20,DBLP:BlanchetLZZ23,DBLP:ZhangLMY0WL23} that used pessimistic reward-based training policies,
we integrate the reliability values into the pessimistic MDP framework.
Specifically, the $\alpha$-pessimistic MDP is defined by the tuple $ \hat{M}_p = \{S, A, \hat{R}, \hat{P}_p, \rho_p, \gamma\} $, where $S$,$A$, $\rho_p$ and  $\gamma$ retain the same definitions as in the original MDP (see Sec.~\ref{sec:preli}).
The transition function and reward function are dynamically updated and evaluated using the truncation metric, defined as follows:

\begin{equation}
    \hat{P}(\cdot|s',\tau_{<t})=\left\{
    \begin{array}{lcl} 
         {0 \quad if\ U_t=1} \\
         {\hat{P}(\cdot|s_t,\tau_{<t}) \quad otherwise},
    \end{array} \right.
\end{equation}

\begin{equation}
\label{eq:a_mdp}
    \hat{R}(s_t,a_t)=\left\{
    \begin{array}{lcl}
         0 \quad if \ U_t=1 \\
         r(s_t,a_t)- \beta \frac{\Gamma(s_t,\tau_{<t})}{\alpha}) \quad otherwise, 
    \end{array}\right.
\end{equation}
\noindent where $\beta$ is the reliability penalty hyperparameters for each state-action pair $(s, a)$.

By introducing the $\alpha$-pessimistic MDP, we incorporate trajectory generation errors into the reward function, ensuring that trajectories with larger errors receive smaller reward values. In this way, the policy learns a lower Q-value when encountering such trajectories, thereby becoming more conservative and robust
~\cite{DBLP:KidambiRNJ20,DBLP:KumarZTL20}.
Moreover, If the truncation metric $U_t = 1$, indicating that the cumulative reliability $\Gamma$ at step $t$  exceeds the threshold $\alpha$, the trajectory generation process will be terminated.

In order to analyze the effect of integrating truncation metric into pessimistic MDP, we assume that $\tilde{r}$ is the maximum reward value in $\alpha$-pessimistic MDP, and then derive the performance bound between the $\alpha$-pessimistic MDP policy and the real MDP policy.


\begin{thm}
	Let $\pi$ be the policy learned from the $\alpha$-pessimistic MDP, and $\pi^*$ be the optimal policy in the true MDP M. Then there exist constants $C$ and $\beta$ such that the performance difference between the two policies in the true MDP satisfies: $\bigl|V^P(\pi^*) - V^P(\pi)\bigr| \;\le\; C\;\alpha\,\beta$, where $V^P(\pi)$ denotes the value function of the policy $\pi$ in the real M.
	\label{th:lower}
\end{thm}

Theorem~\ref{th:lower} demonstrates that incorporating cumulative reliable values into an \(\alpha\)-pessimistic MDP can provide a strict performance lower bound for a policy in the real MDP.
Simply put, by properly setting the hyperparameters $\alpha$ and $\beta$, the performance gap between the learned policy and the true optimal policy can be controlled within the $O(\alpha\beta)$.

However, the cumulative reliability and threshold $\alpha$ cannot be directly derived through the calculation of $D_{IST}$ because $P(\cdot|s,\tau_{<t})$ are typically unknown. 
To solve this issue, we employ a Variational Autoencoder (VAE)~\cite{DBLP:Li024,DBLP:ZhaoZFHSC24} to calculate the distribution error of $(s,a,s')$ as a replacement for the total variation distance  $D_{IST}(P(\cdot | s, a), \hat{P}(\cdot | s, a))$.
Specifically, we use a neural network to learn an encoder and a decoder. 
The encoder maps $x = (s, a, s')$ to a latent vector $z$, while the decoder reconstructs $x$ from $z$.
The conditional probability of the encoder is $ q(z | x)$ , and the constrained probability distribution of the decoder is $p(x | z)$. Then, the loss function of the VAE is expressed as:
\begin{equation}
    L(\psi,\phi)=-\mathbb{E}_{z\sim q(z|x)}[\log(p(x|z))]+D_{\text{KL}}(q(z|x)||p(z)),
\end{equation}
where $\psi$ is the encoder parameter, $\phi$ is the decoder parameter and $D_{\text{KL}}$ is KL divergence.
By learning the VAE, we identify the maximum reconstruction error in the offline data and set it as the threshold $\alpha$.
Next, the generated states and actions use the VAE to calculate the $D_{IST}$, which is then used as a reliable value for the generated data to compute the cumulative reliability of the trajectory.

\subsection{High-return Trajectory Generation  }

While the above method can generate more reliable trajectories, it does not ensure the generation of high-return trajectories.
To address this issue,  we use the reward at time $t$ as a condition to guide the model to generate the action at $t+1$, which is similar to the discriminator-guided generation method used in language models~\cite{DBLP:0011LH024}.
Specifically, RT applies a binary classifier $\mathbb{P}(H_t |\cdots)$ to identify whether an action brings a high reward before taking an action at time $t$, and applies the Bayes' rule to approximate the high reward distribution $\mathbb{P}(R_t, \ldots | H_t) \propto \mathbb{P}(H_t | R_t, \ldots) \mathbb{P}(R_t, \ldots)$, where $H_t$ denotes the high reward at time $t$.
This probabilistic modeling~\cite{DBLP:LeeNYLFGFXJMM22} forms a simple autoregressive process in which $R_t$ is first generated based on a high and reasonable log-probability, and actions are then generated according to $P_\theta(a_t \mid R_t, \ldots)$, which can be considered a variant of the class-conditional model~\cite{DBLP:abs-1909-05858} that automatically takes actions with high rewards at each time step to generate high-return trajectories.

To further enhance the generation of high-return trajectories, RT employs a backward generation mechanism to generate trajectories by randomly selecting a state from the trajectory as the starting point and modeling the states, actions, and cumulative rewards in a backward sequence, thus ensures that the generated trajectories include the actual goal state\footnote{Within the scope of this research, the goal state can be defined as any state along the trajectory that yields a maximal reward.} typically associated with high rewards.
Specifically, each trajectory $\tau_{back}=(s_T,a_T,r_T,R_T,\cdots,s_1,a_1,r_1,R_1)$ is used as an input sequence for RT, which is expressed as follows: 
\begin{equation}
\begin{split}
     \log P_\theta(\tau_t|\tau_{>t}) &= \log P_\theta(s_t|\tau_{>t})+\log P_\theta(a_{t}|s_t,\tau_{>t}) \\
    & +\log P_\theta(r_t|a_{t},s_t,\tau_{>t}) \\
    &+\log P_\theta(R_t|r_t,a_t,s_t,\tau_{>t}),
\end{split}
\end{equation}
where $\tau$ denotes $\tau_{back}$, and $\tau_{>t}$ represents the trajectory from time step $t$ to the terminal state.
The training objective is to maximize:
\begin{equation}
\label{eq:RT_loss}
    \mathcal{L}_\tau=\sum_{t=1}^{T} \log P_\theta(\tau_{T-t+1} | \tau_{> T-t+1}).
\end{equation}

After the training process, the learned RT generates trajectories to augment the offline dataset $D$. For the original trajectories $\tau \in D $ in the original dataset, RT begins generating trajectories backward from time step $T'$:
\begin{equation}
    \begin{split}
        \hat{\tau} &=(\cdots,\hat{s}_{T-T'-1},\hat{a}_{T-T'-1},\hat{r}_{T-T'-1},\hat{R}_{T-T'-1}         \\&s_{T'},a_{T'},r_{T'},R_{T'},\cdots) \sim P_\theta(\tau_{<T-T'}|\tau_{\geq T-T'}),
    \end{split}
\end{equation}
where $\tau_{\geq T-T'}$ denotes the original trajectory truncated at time step $T-T'$.
Then, the generated sequence is concatenated with the truncated original trajectory, resulting in the formation of a new trajectory $\tau'=\hat{\tau}_{<T-T'}+\tau_{\geq T-T'}$, where $+$ is the concatenation operation. 
By adopting this approach, we are able to augment a solitary high-return trajectory into the trajectories, consequently enriching the offline dataset.

\section{Experiments}
In this section, we first compare RT with existing mainstream offline RL methods across various domains in the D4RL benchmark~\cite{DBLP:abs-2004-07219}, including Maze2D, MuJoCo and  AntMaze, and then analyze the impact of different mechanisms or components of RT on its performance. Refer to the Appendix for more details about the environments.

\subsection{Comparison Methods}
Since our goal is to augment offline datasets to improve the performance of offline RL algorithms, we compare RT with recent augmentation algorithms, including \textbf{ROMI}~\cite{DBLP:WangLJZLZ21}, \textbf{CABI}~\cite{DBLP:LyuLL22}),
\textbf{TATU}~\cite{DBLP:ZhangLMY0WL23} and \textbf{DStitch}~\cite{DBLP:LiSZL024}.
In addition, in order to intuitively understand the performance difference between RT and notable offline RL algorithms, RT is compared to the following notable methods, including offline model-free algorithms (i.e., \textbf{BCQ}~\cite{DBLP:FujimotoMP19}, \textbf{CQL}~\cite{DBLP:KumarZTL20}, \textbf{IQL}~\cite{DBLP:KostrikovNL22}, and \textbf{DT}~\cite{DBLP:DT}), as well as offline model-based algorithms (i.e., \textbf{MOPO}~\cite{DBLP:YuTYEZLFM20}, \textbf{MOReL}~\cite{DBLP:KidambiRNJ20} and \textbf{MOPP}~\cite{DBLP:ZhanZX22}.
More description of the above methods is provided in the Appendix.

\begin{figure*}[t]
    \centering
    \subfigure[BCQ]{
        \includegraphics[width=0.31\textwidth]{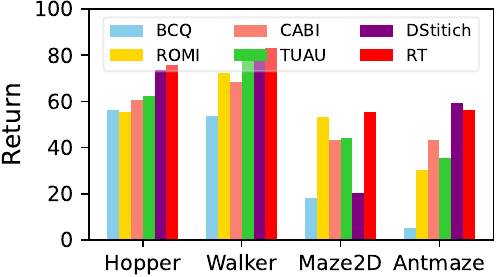}
    }
    \subfigure[CQL]{
        \includegraphics[width=0.31\textwidth]{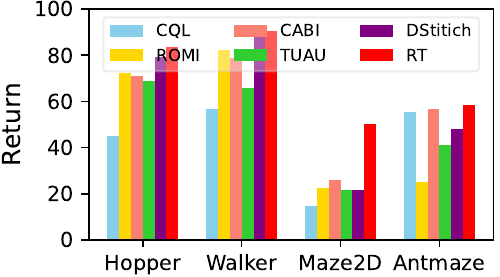}
    }
    \subfigure[IQL]{
        \includegraphics[width=0.31\textwidth]{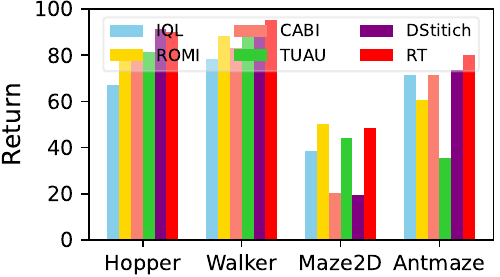}
    }
    \caption{Performance of RT, ROMI, CABI, TATU and DStitch combined with different model-free algorithms (averaged over 10 random seeds).
    }
    
    \label{fig:model-free_boost}
\end{figure*}

\subsection{Results}
\noindent
\textbf{Comparison to augment methods}.
ROMI, CABI, TATU, DStitch and RT are combined with model-free methods (e.g., BCQ, CQL and IQL) and evaluated on Hopper-medium, Walker-medium, sparse Maze2D-medium and diverse Antmaze-medium.
The experimental results shown in Fig.~\ref{fig:model-free_boost} demonstrate that RT consistently outperforms ROMI  CABI and TATU. 
Moreover, both ROMI, CABI and TATU perform worse than IQL alone in the Antmaze-medium environment because their inability to generate high-return trajectories introduces a large amount of unrewarded data, leading to inaccurate policy constraints and then a suboptimal policy.
Although DStitch can merge high-reward and low-reward trajectories, the presence of a high proportion of suboptimal data poses a significant challenge to its ability to find suitable trajectories for merging, as in the Maze2D environment.
In contrast, RT employs sequential modeling to generate high-return trajectories starting from the goal state and utilizes reliability estimation to automatically determine the length of generated trajectories, resulting in more robust performance across different environments.

\begin{figure}[t]
    \centering
    \subfigure[Walker]{
        \includegraphics[width=0.22\textwidth]{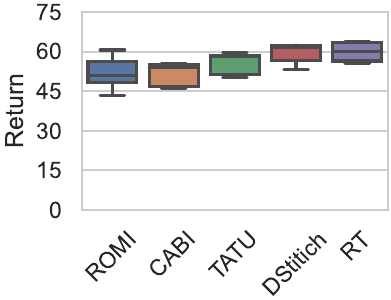}
    }
    \subfigure[Halfcheetah]{
        \includegraphics[width=0.22\textwidth]{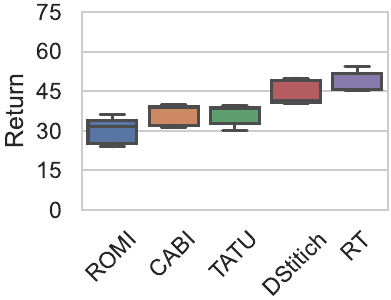}
    }
    \subfigure[AntMaze]{
        \includegraphics[width=0.22\textwidth]{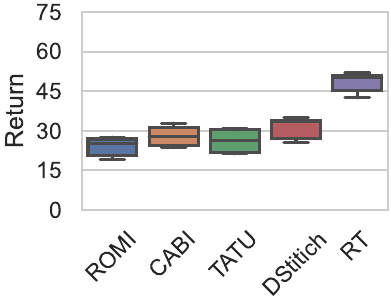}
    }
    \subfigure[Maze2D]{
        \includegraphics[width=0.22\textwidth]{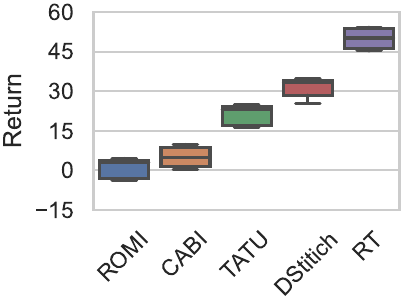}
    }
    \caption{Results of BC learned using trajectories generated by RT, ROMI, CABI, TATU and DStitch (averaged over 10 random seeds). 
    }
    \label{fig:BC_compar}
\end{figure}

Next, to evaluate the return of the trajectories generated by RT, we employ Behavioral Cloning (BC)~\cite{DBLP:TorabiWS18} to learn a policy from the generated data, as the effectiveness of BC is highly dependent on the quality and quantity of the high-return data.
We train ROMI, CABI, TATU, DStitch and RT on the Walker-medium, Halfcheetah-medium, Antmaze-umaze, and Maze2D-umaze environments, and then select 50 trajectories from these environments to generate new trajectories by sampling each selected trajectory 10 times for learning BC.
The results in Fig.~\ref{fig:BC_compar} show that RT shows strong learning performance, while ROMI, CABI, TATU and DStitch exhibit large performance fluctuations in different environments, which can be attributed to the fact that ROMI, CABI, TATU and DStitch prioritize generating trajectories that adhere to the original data distribution without considering the returns of the generated trajectories, preventing BC from learning an effective policy from the generated trajectories.
In contrast, RT takes into account both the reliability and high-return of the data when generating trajectories, which allows to achieve better learning performance across different environments.

\begin{figure}[]
    \centering
    \subfigure[States distribution]{
        \includegraphics[width=0.187\textwidth]{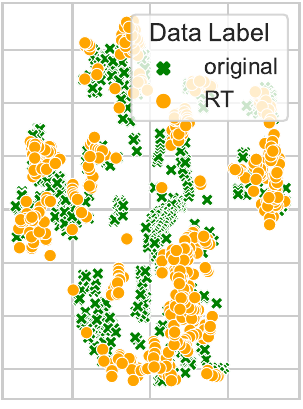}
    }
    \subfigure[States return]{
        \includegraphics[width=0.25\textwidth]{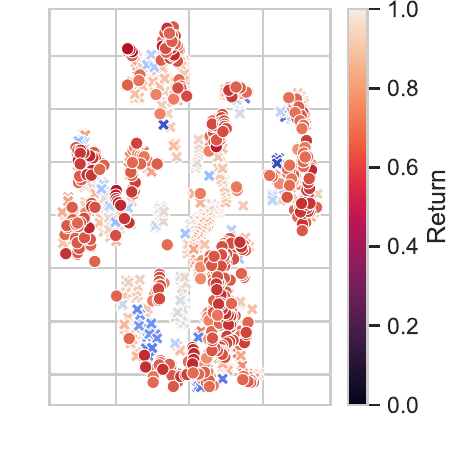}
    }
    \caption{Visualization of generated data. (a) represents the distribution of the original states and the states generated by RT in Walker. (b) represents the cumulative return assessed using the state-value function, with brighter colors indicating higher returns.}
    \label{fig:state_distribution}
\end{figure}

Furthermore, we employe T-SNE~\cite{van2008visualizing,DBLP:ChenWZDLL24} visualization to compare the distribution of states generated by RT with the original states, and evaluate the rewards of each state using a value function.
As shown in Fig.~\ref{fig:state_distribution}, the distribution of samples generated by RT is essentially consistent with that of the original data.
However, when looking at the non-overlapping regions of the generated and original states, we find that the generated data has much higher values than the original state. 
Those out-of-distribution and high-value regions reflect the ability of RT to generate high-return trajectories (refer to the Appendix for results of other environments).
\\

\begin{table*}[]
\centering
\begin{tabular}{llcccccccc}
\hline
              & Environment                                                                            & \multicolumn{1}{l}{BCQ}                                    & \multicolumn{1}{l}{CQL}                                    & \multicolumn{1}{l}{IQL}                                    & \multicolumn{1}{l}{DT}                                     & \multicolumn{1}{l}{MOPO}                                   & \multicolumn{1}{l}{MOReL}                                  & \multicolumn{1}{l}{MOPP}                                   & \multicolumn{1}{l}{RT+BCQ}                                             \\ \hline
medium-replay & \begin{tabular}[c]{@{}l@{}}Hopper\\ Walker\\ Halfcheetah\end{tabular}                  & \begin{tabular}[c]{@{}c@{}}33.3\\ 16.8\\ 39.2\end{tabular} & \begin{tabular}[c]{@{}c@{}}30.6\\ 15.8\\ 40.7\end{tabular} & \begin{tabular}[c]{@{}c@{}}96.8\\ \textbf{74.9}\\ 45.1\end{tabular} & \begin{tabular}[c]{@{}c@{}}83.6\\ 67.2\\ 40.3\end{tabular} & \begin{tabular}[c]{@{}c@{}}68.5\\ 40.3\\ 53.2\end{tabular} & \begin{tabular}[c]{@{}c@{}}93.8\\ 48.7\\ 39.8\end{tabular} & \begin{tabular}[c]{@{}c@{}}35.2\\ 23.6\\ 43.8\end{tabular} & \begin{tabular}[c]{@{}c@{}}\textbf{99.3$\pm$7.2}\\ 70.3$\pm$5.8\\ \textbf{47$\pm$3.5}\end{tabular}   \\ \hline
sparse        & \begin{tabular}[c]{@{}l@{}}Maze2D-umaze\\ Maze2D-medium\\ Maze2D-large\end{tabular}    & \begin{tabular}[c]{@{}c@{}}49.3\\ 18.2\\ 32.6\end{tabular} & \begin{tabular}[c]{@{}c@{}}19.1\\ 14.6\\ 17.2\end{tabular} & \begin{tabular}[c]{@{}c@{}}20.1\\ 10.6\\ 19.3\end{tabular} & \begin{tabular}[c]{@{}c@{}}52.6\\ 13.2\\ 3.4\end{tabular}  & \begin{tabular}[c]{@{}c@{}}1.2\\ 0.0\\ 0.9\end{tabular}    & \begin{tabular}[c]{@{}c@{}}0.0\\ 1.3\\ 0.0\end{tabular}    & \begin{tabular}[c]{@{}c@{}}2.1\\ 0.0\\ 0.6\end{tabular}    & \begin{tabular}[c]{@{}c@{}}\textbf{54.3$\pm$5.6}\\ \textbf{55.3$\pm$4.8}\\ \textbf{40.8$\pm$5.2}\end{tabular} \\ \hline
diverse       & \begin{tabular}[c]{@{}l@{}}Antmaze-umaze\\ Antmaze-medium\\ Antmaze-large\end{tabular} & \begin{tabular}[c]{@{}c@{}}48.8\\ 5.2\\ 1.9\end{tabular}   & \begin{tabular}[c]{@{}c@{}}83.6\\ 55.4\\ 13.6\end{tabular} & \begin{tabular}[c]{@{}c@{}}63.8\\ \textbf{71.3}\\ 44.2\end{tabular} & \begin{tabular}[c]{@{}c@{}}53.6\\ 44.2\\ 24.8\end{tabular} & \begin{tabular}[c]{@{}c@{}}0.0\\ 0.0\\ 0.0\end{tabular}    & \begin{tabular}[c]{@{}c@{}}0.0\\ 0.6\\ 0.0\end{tabular}    & \begin{tabular}[c]{@{}c@{}}0.0\\ 0.0\\ 0.0\end{tabular}    & \begin{tabular}[c]{@{}c@{}}\textbf{71.3$\pm$4.5}\\ 56.4$\pm$5.4\\ \textbf{46.8$\pm$3.9}\end{tabular} \\ \hline
\end{tabular}
\caption{Normalized returns of different methods on the D4RL benchmark (Bold black font indicates the highest return).}
\label{tab:comp_resut}
\end{table*}

\noindent\textbf{Comparison to notable offline methods}.
Table~\ref{tab:comp_resut} shows the results of RT+BCQ compared with recent offline RL methods (refer to the Appendix for the  full results). 
The experiments demonstrate that RT can significantly enhance the performance of the original algorithm and outperform all the baselines.
It can be seen that the model-based methods have large performance fluctuations in different environments due to the failure in generating high-return trajectories from offline data; for instance, MOReL shows performance comparable to model-free methods in the MuJoCo environment but accumulates almost zero rewards in Maze2D and Antmaze.
RT uses Reliability-guaranteed sequence modeling to adaptively determine trajectory lengths and employs high reward as conditions to generate high-return trajectories from offline data, making it highly effective across diverse environments.


\begin{table}[t]
\begin{tabular}{llccc}
\hline
        & Environment                                                                            & \multicolumn{1}{l}{BCQ}                                    & \multicolumn{1}{l}{FT+BCQ}                                             & \multicolumn{1}{l}{RT+BCQ}                                             \\ \hline
\rotatebox[origin=c]{90}{sparse}  & \begin{tabular}[c]{@{}l@{}}Maze2D-umaze\\ Maze2D-medium\\ Maze2D-large\end{tabular}    & \begin{tabular}[c]{@{}c@{}}49.3\\ 18.2\\ 32.6\end{tabular} & \begin{tabular}[c]{@{}c@{}}50.1$\pm$6.7\\ 42.6$\pm$4.9\\ 33.2$\pm$6.8\end{tabular} & \begin{tabular}[c]{@{}c@{}}\textbf{54.3$\pm$5.6}\\ \textbf{55.3$\pm$4.8}\\ \textbf{40.8$\pm$5.2}\end{tabular} \\ \hline
\rotatebox[origin=c]{90}{dense}   & \begin{tabular}[c]{@{}l@{}}Maze2D-umaze\\ Maze2D-medium\\ Maze2D-large\end{tabular}    & \begin{tabular}[c]{@{}c@{}}50.1\\ 41.3\\ 75.2\end{tabular} & \begin{tabular}[c]{@{}c@{}}55.3$\pm$6.7\\ 60.1$\pm$8.9\\ 83.4$\pm$6.2\end{tabular} & \begin{tabular}[c]{@{}c@{}}\textbf{68.9$\pm$8.5}\\ \textbf{80.6$\pm$8.2}\\ \textbf{98.6$\pm$5.8}\end{tabular} \\ \hline
\rotatebox[origin=c]{90}{fixed}   & Antmaze-umaze                                                                          & 79.5                                                       & 80.3$\pm$7.7                                                               & \textbf{83.7$\pm$8.9}                                                               \\ \hline
\rotatebox[origin=c]{90}{play}    & \begin{tabular}[c]{@{}l@{}}Antmaze-medium\\ Antmaze-large\end{tabular}                 & \begin{tabular}[c]{@{}c@{}}1.3\\ 2.1\end{tabular}          & \begin{tabular}[c]{@{}c@{}}32.2$\pm$6.5\\ 16.2$\pm$8.3\end{tabular}            & \begin{tabular}[c]{@{}c@{}}\textbf{40.6$\pm$6.4}\\ \textbf{23.2$\pm$5.6}\end{tabular}            \\ \hline
\rotatebox[origin=c]{90}{diverse} & \begin{tabular}[c]{@{}l@{}}Antmaze-umaze\\ Antmaze-medium\\ Antmaze-large\end{tabular} & \begin{tabular}[c]{@{}c@{}}48.8\\ 5.2\\ 1.9\end{tabular}   & \begin{tabular}[c]{@{}c@{}}63.5$\pm$4.6\\ 43.4$\pm$8.3\\ 16.6$\pm$9.6\end{tabular} & \begin{tabular}[c]{@{}c@{}}\textbf{71.3$\pm$4.5}\\ \textbf{56.4$\pm$5.4}\\ \textbf{46.8$\pm$3.9}\end{tabular} \\ \hline
\end{tabular}
\caption{Performance comparison between forward and backward generation}
\label{tab:ablation_RFT}
\end{table}

\subsection{Further Analysis}
\noindent\textbf{Generated trajectory analysis}.
To better understand the effectiveness of RT in addressing unreliable trajectory problems, we construct a game called BoxBall, where the ball needs to move from the starting point to the goal state using up, down, left, and right actions while navigating around a wall in front of the goal state.

\begin{figure}[t]
    \centering
    \subfigure[Random]{
        \label{fig:boxball_random}
        \includegraphics[width=0.20\textwidth]{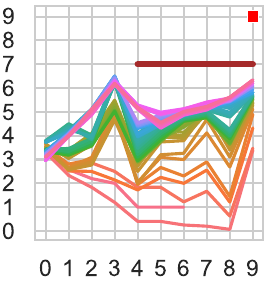}
    }
    \subfigure[Forward]{
        \includegraphics[width=0.20\textwidth]{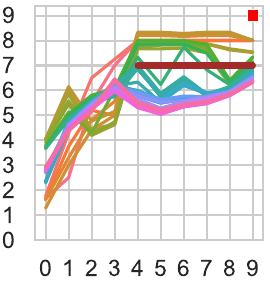}
    }
    \subfigure[Backward]{
        \includegraphics[width=0.20\textwidth]{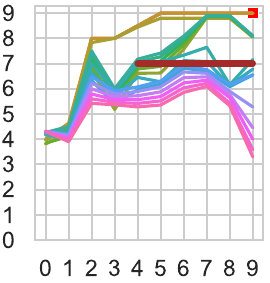}
    }
    \subfigure[RT]{
        \label{fig:boxball_RT}
        \includegraphics[width=0.20\textwidth]{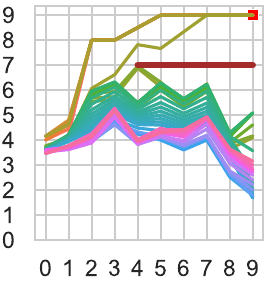}
    }
    \caption{Trajectory Visualization of BoxBall. The moving space of the ball is from 0 to 9, and it moves one grid at a time. The brown line beneath the red square represents the wall in the game. (a)$\sim$(d) denote the visualization trajectories of the random policy, forward model, backward model and RT, respectively.
    }
    \label{fig:boxball_f}
\end{figure}

\begin{figure}[t]
    \centering
    \subfigure[Halfcheetah]{
        \includegraphics[width=0.22\textwidth]{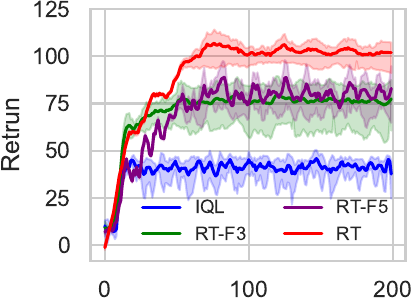}
    }
    \subfigure[Walker]{
        \includegraphics[width=0.22\textwidth]{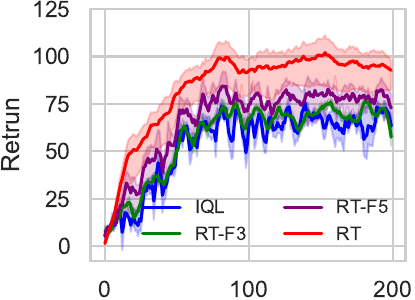}
    }
    \subfigure[Maze2D-large]{
        \includegraphics[width=0.22\textwidth]{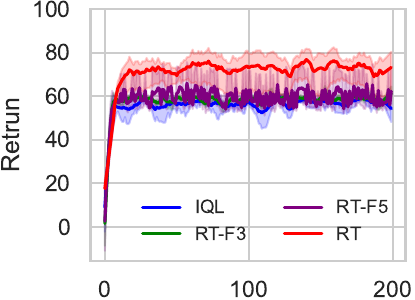}
    }
    \subfigure[AntMaze-large]{
        \includegraphics[width=0.22\textwidth]{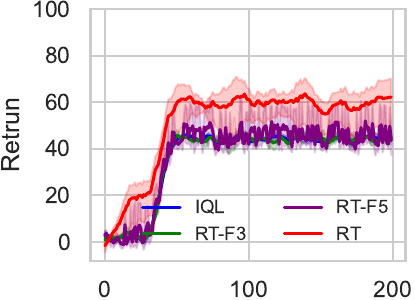}
    }
    \caption{The impact of different truncation mechanisms on the performance of IQL. The x-axis represents the number of training episodes ($1\times 10^4$), and the y-axis represents the cumulative reward.}
    \label{fig:IQL_boost}
\end{figure}

In this environment, we use DQN to collect a dataset and use it to train forward, backward, and RT models.
Then, we use the trained forward model, backward model and RT to generate and visualize 50 trajectories.
As shown in Fig.~\ref{fig:boxball_f}, the trajectory of the random strategy does not include the action of walking through the wall, while the trajectories generated by the forward model contain many unreliable trajectory (i.e., trajectories cross the wall) and fail to reach the goal state.
The values of unreliable trajectories may be arbitrarily misestimated, and then the erroneous values may be propagated throughout the state-action space leading to overestimation of Q-values during training.
Despite employing the same backward trajectory generation method, the backward model generates some unreliable trajectories due to cumulative errors and model lacks of ability to capture long-term dependencies.
RT utilizes sequence modeling techniques to effectively learn the distribution of trajectories and employs reliability estimation to adaptively determine the generated trajectories, which ensures that actions involving passing through walls do not occur.
A more detailed analysis of the generating reliable trajectory is discussed in the Appendix.
\\
\\
\textbf{Forward and backward analysis}.
In order to explore whether backward generation is more advantageous than forward generation,  we conduct forward and backward comparison experiments on Maze2D and Antmaze, where forward generation is donoted as FT.
Table~\ref{tab:ablation_RFT} shows the experimental results for BCQ, FT+BCQ, and RT+BCQ with 10 random seeds. As can be seen, FT+BCQ provides superior performance in most environments, but not as significant as RT+BCQ.
For example, in Antmaze-large, where the maze area is extensive and the paths are complex, FT requires longer generation lengths to generate high-return trajectories, which in turn increases trajectory distribution error.
In contrast, RT is the backward generation, which enables safe generalization across all layouts (i.e., RT can ensure that the generated trajectories contain goal states), thus ensuring RT+BCQ significantly outperforms FT+BCQ and BCQ.
\\
\\
\noindent \textbf{Reliability estimation analysis}.
In order to explore the impact of reliability estimation on RT performance, we conduct experiments on Halfcheetah-medium-replay, Walker-medium-replay, dense Maze2D-larger and diverse Antmaze-large.
RT is divided into two categories: fixed-length generation, with lengths set to 3 (RT-F3) and 5 (RT-F5), and reliability estimation.
Fig.~\ref{fig:IQL_boost} illustrates the impact of different generation length mechanisms on the convergence of the IQL learning process.
Both RT-F3 and RT-F5 can improve the performance of IQL in the Halfcheetah-medium-replay and Walker-medium-replay because of the sequence modeling, which ensures that the data generated by the models are consistent with the original offline data.
RT uses reliability estimation, which makes it more flexible and diverse when generating trajectories, thereby improving the performance of IQL.

In the dense Maze2D-large and diverse Antmaze-large environments, where maze layouts and paths become increasingly complex, shorter trajectories may fail to yield higher cumulative rewards, or may even decrease cumulative rewards.
RT-F3 and RT-F5 generate shorter trajectories by using fixed-length generation to prevent the generated data from exceeding the distribution of offline data, which also limits the generation of high-return trajectories.
In contrast, RT employs a reliability estimation, which can automatically determine the length of the generated trajectories while ensuring that these state-action pairs remain within the distribution range of the original offline data.
Hence, RT can generate more high-return trajectories to facilitate IQL learning.

\subsection{Related Work}
\noindent
Offline model-based methods~\cite{DBLP:KidambiRNJ20,DBLP:SwazinnaUR21} learn an approximate dynamics model of the environment and utilize planning algorithms to search for optimal trajectories within that model.
However, research indicates that even small generation errors can significantly degrade the performance of multi-step rollouts, as generation errors compound, causing the model to deviate from the high-precision region after a few steps~\cite{DBLP:Talvitie17,DBLP:AsadiML18}.
To mitigate generation errors, some studies~\cite{DBLP:YuTYEZLFM20,DBLP:abs-2111-11097,DBLP:ZhengWXH23} employ multiple models to predict states and actions, using uncertainty quantification to eliminate samples with excessive errors.
In addition, some studies~\cite{chen2022lapo,DBLP:YangJZMS22,DBLP:MaT0M23} restrict the learning policy from accessing areas with significant differences between the learned and the real dynamics to prevent policy learning failures due to generation errors.

Recent studies~\cite{DBLP:WangLJZLZ21,DBLP:LyuLL22,DBLP:LuBTP23,DBLP:abs-2406-12550} suggest the use of fixed truncation techniques~\cite{DBLP:JannerFZL19} to generate shorter trajectories and thus reduce the impact of generation errors on generated data.
This generated data is then combined with the original data to create a new dataset, which can be utilized in model-free offline algorithms.
However, these studies neglect the influence of historical information on environmental dynamics, potentially resulting in the generation of data that fails to align with the environment.
Unlike the above methods, RT extends the offline dataset by capturing historical information between different locations in the input sequence and using reliability assessment to generate reliable trajectories.
Moreover, RT employs a cumulative reliability metric to dynamically adjust the trajectory length, enabling a more flexible mitigation of the adverse effects of accumulated generation errors.
\\
\\
\textbf{Sequence modeling}.
Recent research~\cite{DBLP:JannerLL21,DBLP:DT,DBLP:YamagataKS23,DBLP:BadrinathFNB23,DBLP:GaoWCKZ024} shifts the paradigm of RL from traditional MDP to sequence modeling.
In particular, Trajectory Transformer (TT)~\cite{DBLP:JannerLL21} treats offline RL as a sequence modeling problem and further leverages the capabilities of sequence models by using beam search to incorporate states, actions, and rewards.
At the some time, Decision Transformer (DT)~\cite{DBLP:DT} learns the distribution of trajectories and predicts actions based on the given target reward and preceding states, rewards, and actions.
Q-Learning DT~\cite{DBLP:YamagataKS23} extends the use of DT in offline RL by introducing a value function, which relabels the returns in the training data using the dynamics planning results, and then uses the relabeled data to train the DT.
Multi-Game Decision Transformer (MGDT)~\cite{DBLP:LeeNYLFGFXJMM22} trains a Transformer to solve multiple Atari games, which eliminates the need for additional fine-tuning of unseen tasks.
Waypoint Transformer (WT)~\cite{DBLP:BadrinathFNB23} utilizes an architecture constructed based on the DT framework and conditioned on automatically generated waypoints to learn optimal Transformer policies.
Compared to these model-free approaches, we propose a model-based approach that generates high-return trajectories from suboptimal datasets using the Reliability-guaranteed Transformer framework  to facilitate policy learning in model-based RL.

\section{Conclusion}
In this paper, we propose a novel model-based offline RL method RT to facilitate policy learning by generating reliable and high-return trajectories.
RT leverages Reliability-guaranteed sequence generation techniques and conditions the generation process on high rewards to generate high-return trajectories that are legitimate in the environment.
Extensive experiments demonstrate that RT can be effectively integrated with existing model-free algorithms and achieve better performance against existing offline RL benchmarks.
However, RT presupposes a fully observable state space in its empirical evaluations.  
For partially observable environments, RT may require additional techniques (e.g., Kalman Filters) to address the issue of incomplete information in order to accurately predict states and rewards. 
Future work includes addressing this above issue and extending RT to multi-agent offline algorithms.






\bibliographystyle{named}
\bibliography{ijcai25}

\end{document}